# EVALUATION OF GPT-4 FOR CHEST X-RAY IMPRESSION GENERATION: A READER STUDY ON PERFORMANCE AND PERCEPTION

Sebastian Ziegelmayer[1], Alexander W. Marka[1], Nicolas Lenhart[1], Nadja Nehls[1], Stefan Reischl[1], Felix Harder[1], Andreas Sauter[1], Marcus Makowski[1], Markus Graf[1], and Joshua Gawlitza[1]

[1]Department of Diagnostic and Interventional Radiology, School of Medicine Klinikum rechts der Isar, Technical University of Munich, Germany

## ABSTRACT

The remarkable generative capabilities of multimodal foundation models are currently being explored for a variety of applications. Generating radiological impressions is a challenging task that could significantly reduce the workload of radiologists. In our study we explored and analyzed the generative abilities of GPT-4 for Chest X-ray impression generation. To generate and evaluate impressions of chest X-rays based on different input modalities (image, text, text and image), a blinded radiological report was written for 25-cases of the publicly available NIH-dataset. GPT-4 was given image, finding section or both sequentially to generate an input dependent impression. In a blind randomized reading, 4-radiologists rated the impressions based on "Coherence", "Factual Consistency", "Comprehensiveness", and "Medical harmfulness" and were asked to classify the impression origin (Human, AI), providing justification for their decision. Lastly text model evaluation metrics and their correlation with the radiological score (summation of the 4 dimensions) was assessed. According to the radiological score, the human-written impression was rated highest, although not significantly different to text-based impressions. The automated evaluation metrics showed moderate to substantial correlations to the radiological score for the image impressions, however individual scores were highly divergent among inputs, indicating insufficient representation of radiological quality. Detection of AI-generated impressions varied by input and was 61% for text-based impressions. Impressions classified as AI-generated had significantly worse radiological scores even when written by a radiologist, indicating potential bias. Our study revealed significant discrepancies between a radiological assessment and common automatic evaluation metrics depending on the model input. The detection of AI-generated findings is subject to bias that highly rated impressions are perceived as human-written.

## 1 Introduction

In the rapidly evolving landscape of artificial intelligence (AI), foundation models have emerged as a transformative force, particularly for natural language processing. Generative models trained on large-scale datasets, have demonstrated an unprecedented ability to generate human-like text [1], potentially revolutionizing a multitude of sectors. While early models were limited to text, current models allow for multimodal inputs, often image and text pairs, which is promising for the medical sector as clinical decision making is complex and based on a multitude of available data (imaging, blood work, clinical examination etc.). Furthermore, these models show unique behaviours, such as the ability to solve untrained tasks (zero-shot learning) [2] and to deduce correct solutions with few given examples (in-context learning) [3]. Despite the absence of explicit training, these models show an exceptional capacity for clinical reasoning. For example, GPT-4 [4] passed the United States Medical Licensing Examination (USMLE), outperforming predecessor models and large language models (LLM) that were explicitly fine-tuned for the medical sector [5]. In medical imaging, the applications are manifold, and it has been shown that models can not only draw radiological conclusions [6], but also structure reports [7] and even generate impressions or reports based on the findings given in a report [8] or the image itself [9]. Writing image impressions is a complex core task of radiologist and requires a comprehensive and comprehensible summation and interpretation of the findings, with appropriate recommendations. It is the decisive passage in a radiological report for the appropriate care of the patient [10, 11]. One of the leading obstacles limiting the development of models for generating clinically applicable reports is the lack of evaluation metrics that capture the core aspects of radiological impressions, and new scores are currently being developed [12, 13]. While there are initial studies on the perception of AI-generated text in the general population [14] ,insights are missing for specialized areas such as medical imaging. Therefore our study investigated the ability of GPT-4 to generate radiological impressions based on different inputs (Image, Text, Text and Image), focusing on the correlation between radiological assessment of impression quality and common automated evaluation metrics, as well as radiological perception of AI-generated text.

## 2 Methods

### 2.1 Dataset and Report generation

25 cases were randomly selected from the publicly available NIH Chest X-ray dataset, consisting of 100,000 de-identified images of chest radiographs with various pathologies [15]. Due to the publicly available dataset used in this study, the requirement to obtain written informed consent from the subjects was waived. For each case a full report with "findings" and "impression" section was written by a blinded board-certified radiologist. Based on the outstanding "out-of-box" performance of GPT-4 and to allow for future comparison prompt engineering, for example retrieval augmented generation, chain-of-thought prompting was reduced to a minimum. The only augmentation was done by including the word sequence "Let's think step by step" in the prompt as it was shown to improve the zero shot capabilities of the model [16]. For each case the chest x-ray image, the finding section from the written report and a combination of both was used as input. The following fixed prompt structure was used for all inputs {Imagine you are a radiologist. Generate a short radiological impression based on the main findings in (…). Use medical vocabulary. Let's think step by step.}. For each input the parentheses were specified as "text", "image" and "text and image".

### 2.2 Radiological reading

A questionnaire adapted from Sun and Ong et al. was prepared and extended with the questions "Is the impression written by a human or an AI-model" and "What is the reason you think this is an AI-generated impression" [8]. The original questionnaire included four dimensions "Coherence", "Factual consistency", "Comprehensiveness" and "Medical harmfulness", which were used to derive a radiological score based on a 5-point Likert scale for each dimension. Participants were given a short explanation for each dimension. For "Factual consistency" and "Medical harmfulness", a follow-up question was asked to specify the reasons why participants thought the impression was factually inconsistent and harmful. Each dimension was equally weighted resulting in a maximum score of 20 points and a lowest score of 4 points. The radiological score was recorded for each impression. Four radiologists from our institution (avg. experience in years = 7.7, range 3-12 years) rated the impressions based on the X-ray image and the finding section. The reading was done anonymously and blinded. Readers were presented with 4 randomly ordered impressions per case; the distribution of AI-generated and human-written impression was not known to the readers. For the question: "What is the reason you think this is an AI-generated impression" four choices were presented ("Error/Factual consistency", "Spelling/Grammar", "Structure/Coherence", "Comprehensiveness") readers were also able to give other reasons if the choices were not applicable.

### 2.3 Extraction of model metrics for text evaluation

The extraction of the automatic metric to evaluate the generated impressions were based on the study of Yu, Endo and Krishan et al. [13]. The following metrics were extracted:BLEU [17], BERT [18], CheXbert vector similarity (CheXbert semb) [19], RadGraph [12] and the composite metric RadCliQ (Radiology Report Clinical Quality) [13], which combines BLEU and Radgraph. The BLEU score is calculated based on the number of matching n-grams (contiguous sequences of n words) in the predicted and reference texts. The score includes a penalty for predicted texts that are shorter than the references (brevity penalty). BERT score additionally tries to leverage contextual embeddings to include text semantics. Both the candidate (generated) text and the reference text are tokenized and fed into a pre-trained BERT model to obtain contextual embeddings for each token, which are then compared by calculating cosine-similarity and maximum similarity. CheXbert vector similarity and RadGraph are two scores that were developed to specifically evaluate radiological reports. CheXbert vector similarity defines a reference vector with the 13 most common labels (findings in a X-ray) and a no finding label and measures the cosine similarity. The Radgraph metric converts the texts into a knowledge graph with radiological dependencies and semantics and measures the overlap for both the generated and the ground truth text. In accordance to Yu, Endo and Krishan et al [13] the used model checkpoint and the inference code can be found at [20] and [21] respectively. Lastly the proposed RadCliQ composition metric was extracted, which combines BLEU and Radgraph F1 (the harmonic mean of precision and recall) and is regarded as an estimate for the radiological error rate.

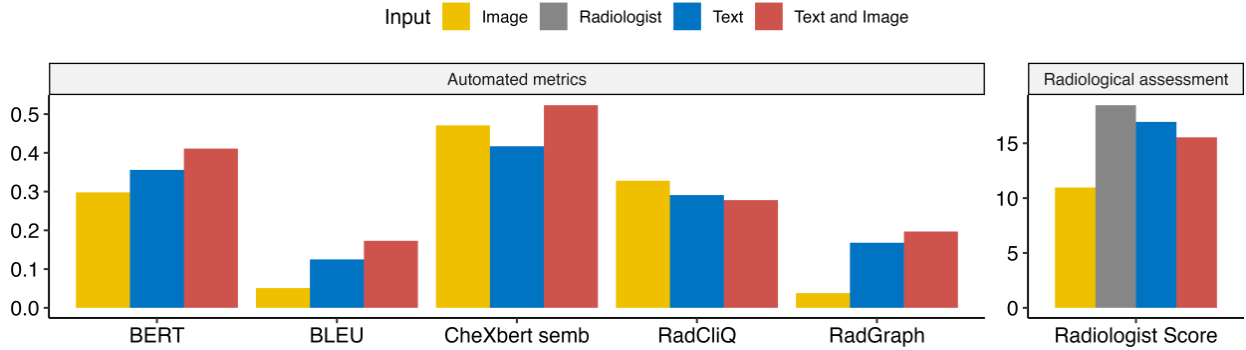

Figure 1: Barplots for each score depending on the respective input (Image = Blue, Text = Yellow, Text and Image = Red, Radiologist = Gray). All model metrics are categorized as "Automated metrics". RadCliQ was scaled by 10-1 to improve visual comparability. Except for RadCliQ which correspond to the error rate, a higher score indicates a better approximation. For the automated metrics, the text and image-based impression score was highest, while the radiological score for the text-based impression was closest to the human ground truth.

## 2.4 Statistical modelling

All statistical modeling was done in Python and R. To evaluate the correlation between the automatic evaluation metrics and the radiological score, Kendall rank correlation coefficients were calculated. To compare each evaluation metric based on the input one-way ANOVA was calculated for all possible combinations. Significant ANOVA were followed up by Tukey post-hoc tests to perform multiple pairwise comparisons. Students t-test was used to compare the radiological scores between AI and human classified impressions. For all statistical tests a p-value below 0.05 was deemed significant.

# 3 Results

## 3.1 Evaluation of generated and written Impressions

For each score the impressions were analysed based on the input using ANOVA and Tukey post-hoc tests if the latter was significant. Except for CheXbert vector similarity, all scores showed significant differences in the ANOVA. For the automatic evaluation metrics the impression based on the combined text and image achieved the highest score, although not significantly higher than the text impressions (Turkey post-hoc tests text and image vs text: BLEU: difference=0.04 p=0.102, BERT: difference=0.05 p=0.31, CheXbert vector similarity: difference=0.11 p=0.27, RadGraph: difference=0.03 p=0.61, RadCliQ: difference=0.14 p=0.26). Image impressions were generally outperformed by text and text and image impressions, except for the CheXbert vector similarity (Turkey post-hoc tests image vs text/text and image: BLEU: difference=0.07/0.12 p≤ 0.005/0.0001, BERT: difference=0.06/0.11 p=0.21/0.006, CheXbert vector similarity: difference=-0.50/0.10 p=0.71/0.27, RadGraph: difference=0.13/016 p≤ 0.0005/0.0001, RadCliQ: difference=-0.36/-0.50 p≤ 0.0005/0.0001). Based on the radiological score the written impression by the radiologist achieved the highest score, although not significantly higher than text impressions (Turkey post-hoc tests radiologist vs text/text and image/image: difference= 1.52/2.93/7.50 p=0.23/0.005/0.0001). Score differences for all inputs is shown in Figure1.

## 3.2 Metric correlation between automatic and radiological evaluation

The correlation between the radiological score and the automated metrics was calculated for all inputs. For the image impressions, a significant moderate correlation was shown for all scores, whereas for text and text and image impressions only single scores correlated significantly (Text - RadGraph, Text and Image - CheXbert vector similarity). Strong variations were noted for the individual metrics depending on the input (exemplary RadGraph image impression R:0.47, text and image impression R:-0.25), except for CheXbert vector similarity, which showed a positive correlation for all inputs, although not significant for text input. All correlation coefficients between the radiological score and the automated metrics are summarized in Table1 and visualized for all inputs in Figure2 .

Table 1: Correlation coefficients for each input and metric with the respective p-value

|  | Image | P | Text | P | Text and Image | P |
|---|---|---|---|---|---|---|
| **BLEU** | 0.29 | .042 | -0.18 | .210 | 0.027 | .850 |
| **BERT** | 0.37 | .010 | -0.093 | .530 | 0.3 | .041 |
| **CheXbert vector similarity** | 0.19 | .019 | 0.1 | .500 | 0.49 | <.001 |
| **RadGraph** | 0.47 | .003 | -0.18 | .210 | -0.25 | .091 |
| **RadCliQ** | -0.41 | .004 | 0.18 | .210 | 0.17 | .250 |

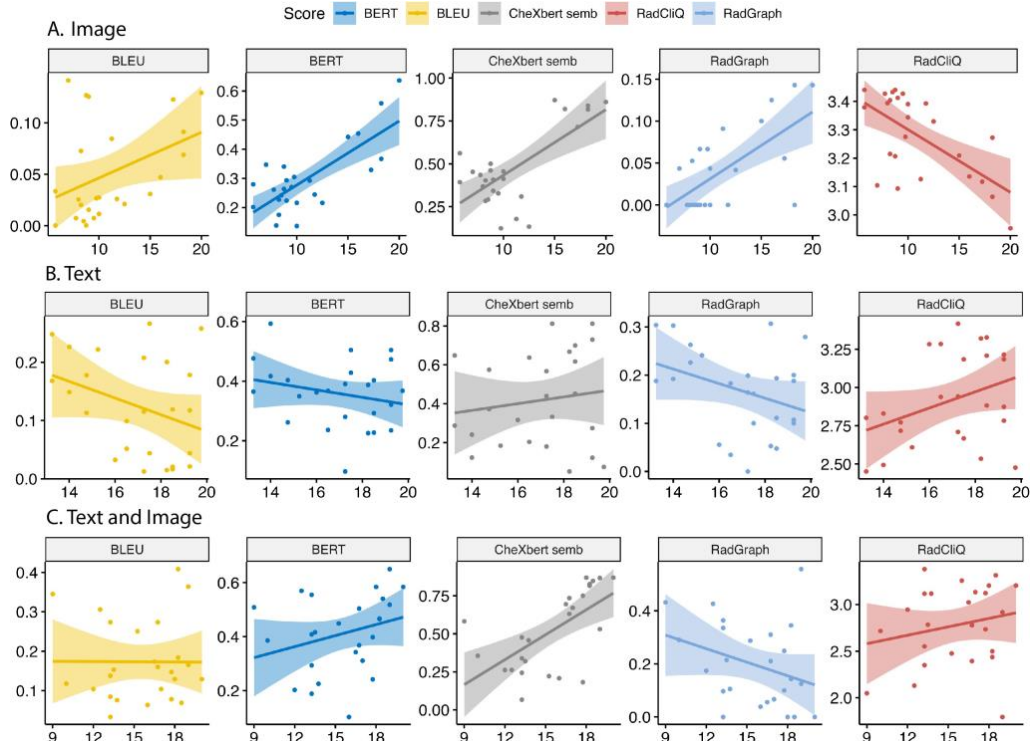

Figure 2: Scatterplots for each automated metric (BERT = Blue, BLEU = Yellow, CheXbert vector similarity = Gray, RadGraph = Lightblue, RadCliQ = Red) depending on the input (A. Image, B. Text, C. Text and Image). For the image input all metrics showed significant correlation. However, correlation was divergent to opposing for the text and text and image inputs.

### 3.3 Radiological perception of AI-generated impression

Four radiologists were asked to classify each impression as AI-generated or human-written and justify their decision if AI-generated was chosen. The accuracy rates to distinguish between human-written and AI-generated impressions diverged substantially based on the underlying input with the highest accuracy for image input and the radiologist written impressions of 87%. Accuracy rates were substantially lower for text and text and image input with only 61% and 63% classified correctly as AI-generated. All detection rates are shown in Figure3.. An analysis of justification for classification as AI-generated showed contrasting results for the different inputs. Image-based impressions were classified as AI-generated with 85% due to Factual consistency/Error, whereas text and text and image showed a more homogeneous distribution of reasons. In particular, text-based impressions showed a similar distribution of justifications as human-written assessments falsely classified as AI generated (Radiologist Impression/Text Impression: Comprehensiveness: 7.7%/8.2%; Error/Factual consistency: 46.1%/31.1%; Spelling/Grammar: 23.1%/26.2%; Structure/Coherence: 23.1%/34.4%). The distribution for each input is shown in the mosaic plot in Figure4. Lastly correlation between the radiological score and the binary impression origin was investigated. It was shown that impressions classified as human-written were significantly higher rated by the radiologist, with a mean score of 18.11 for impressions classified as human written and 13.41 ($p \leq 0.0001$) for impressions classified as AI generated.

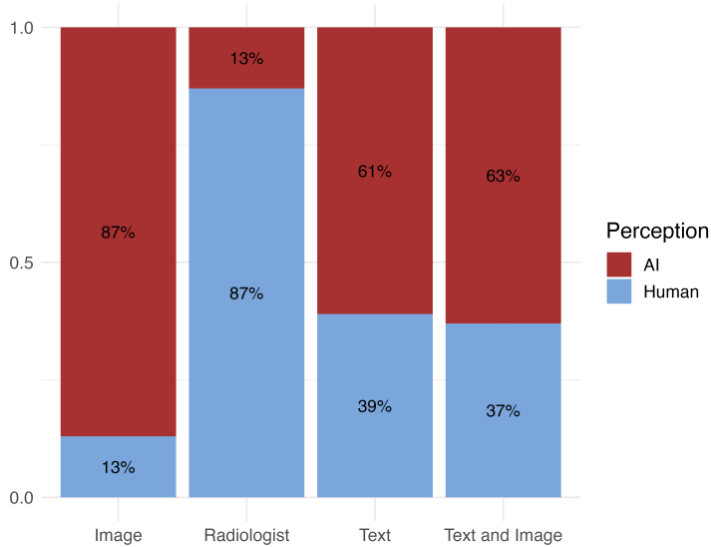

Figure 3: Barplots showing the classification percentages of the impression origin for each input and all readers (AI-generated = Red, Human-written = Lightblue).

## 4 Discussion

In our study we investigated the zero-shot capabilities of GPT-4 to generate radiological impressions of chest X-rays from image and text input. Impressions were analysed using current model evaluation metrics for generated text and a radiological scoring system. The latter showed that human-written impressions achieved the highest score, however text-based impressions were not rated significantly lower. The correlation between automated metrics and the radiological score showed strong deviations for the different inputs. The radiological detection rate of AI-generated impressions was dependent on the input and was found to be only 61% for text-based impressions. Impressions that were classified as human-written, independent of their origin, received a significantly higher radiological score than impressions that were classified as AI-generated.

Generative Models trained on large-scale corpora have shown clinical reasoning without being specifically aligned [5], particularly GPT-4 has shown remarkable capabilities comparable to state-of-the-art foundation models fine tuned on medical data [22]. We evaluated its "out-of-the-box" performance for chest X-ray impression generation based on different inputs. Based on the radiological score text-based impressions were not rated significantly lower than the radiological impressions (difference 1.53, p=0.23), whereas image and text and image-based impressions were rated significantly lower. We followed the rating system of Sun and Ong et al. [8] who showed that text-based impressions rated by radiologists were inferior, but comparable when rated by non-radiologists, indicating inter-reader variability. However, the study did not clarify if the radiological evaluations of the impressions were conducted under blinded conditions. Our work identified radiological bias as impressions classified as human-written, received higher ratings. Without blinding the reader, there is a risk that the inferiority of the AI-generated impressions is due to bias.

In contrast to the radiological score, for the automated metrics, the impressions based on text and image was rated the closest to the radiological impressions, followed by text-based impressions except. For the image-based impressions, there was a significant moderate correlation between the automated metrics and the radiological score, however, for the other inputs opposite or non-significant correlations were found. Automatic metrics that capture relevant aspects of radiological quality are a prerequisite for the successful development and clinical integration of such models and the focus of current research. It has been shown that metrics used for general LLMs doesn't represent radiological quality of a report [23] and developments of evaluation metrics that explicitly approximate radiological quality (RadGraph, RadCliQ) led to an improvement in model development [24]. Although the latter also showed deviations depending on the input, the significance of the radiological scores needs to be discussed. Evaluation metrics can only be as good as the human assessment, which is not free of bias and characterized by false heuristics [14]. Our findings underline this point, as impressions that were classified as human written scored were significantly scored significantly higher by the radiologist. Additionally radiological written as well as text generated impressions, show a similar justification for being classified as AI-generated, which indicates false perceptions for AI-generated language. The writing of a radiological report is complex and the core aspects representing radiological quality need to be further specified [10].

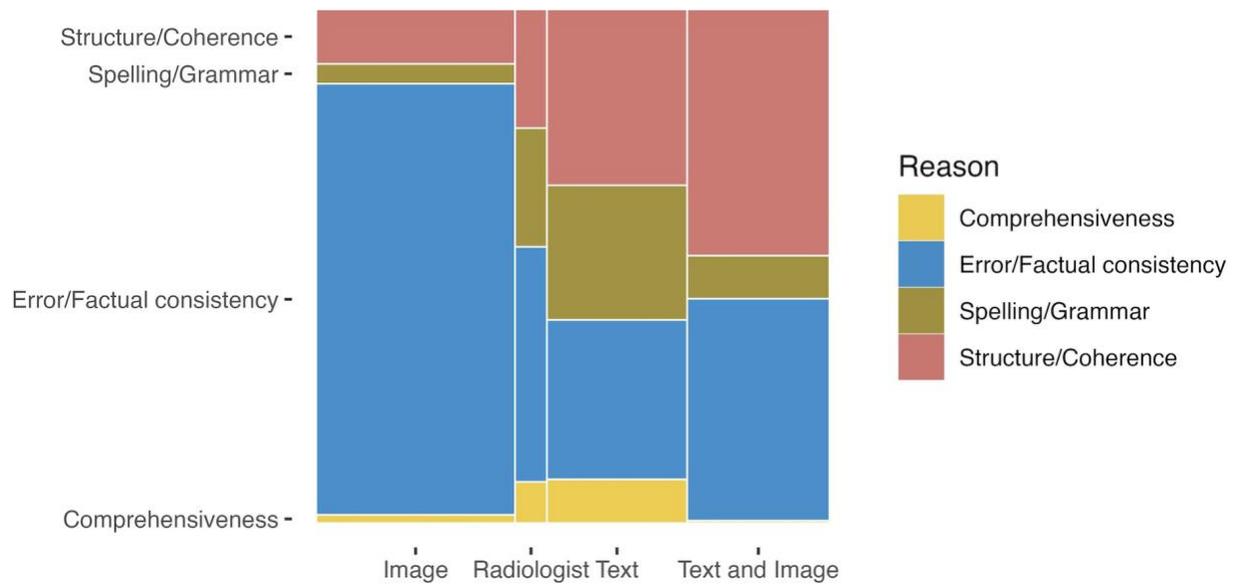

Figure 4: Mosaic-plot visualizing the justification for classifying an impression as AI-generated with the true origin on the x-Axis and the reasons on the y-Axis. The width of each bar corresponds to the frequency of being classified as AI-generated. The height of each colored bar corresponds to the percentage within the given origin.

Our study has several limitations that need to be addressed. The small sample size may not capture the variety of disease represented in chest X-rays and all aspects and nuances of radiological impression writing. However, the correlation of the metrics as well as the radiological perception, which were the focus of the study, are only influenced by this to a limited extent. Not all available metrics for assessing generated text have been explored. However, due to the rapid development of this research field, care was taken to include the most current and promising metrics [13]. Finally, the radiological score was not free of bias and interrater variability is a common problem not only in radiological reporting [25], but in human assessment in general [26]. We have tried to quantify the evaluation of radiological quality by means of 4 dimensions. However, it was discovered that there is a bias that findings are higher rated if they are perceived as humanly written. Human evaluation is not error-free, but it is the benchmark for the evaluation of generated text. Therefore, radiological heuristics, sources of error and relevant aspects of radiological quality should be further investigated as they are essential for the development of useful model metrics.

## 5 Conclusion

In summary, our study investigates the correlation of radiological assessments with automatic evaluation metrics for multimodal chest X-ray impression generation. Based on a radiological score text-based generations were not inferior to human written impressions. Depending on the input, strong deviations of the radiological assessments and the automated metrics were shown, however radiological evaluation is not free of bias.